\documentclass[twoside]{article}
\usepackage[accepted]{aistats2e}
\usepackage{amsmath}
\usepackage{amsthm}
\usepackage{amssymb}
\usepackage{algorithm,algorithmic}
\usepackage{graphicx}
\usepackage{subfigure}
\newtheorem{theorem}{Theorem}

\newtheorem{prop}{Proposition}
\newtheorem{defn}{Definition}

\begin{document}



\twocolumn[

\aistatstitle{Learning Planar Ising Models}
% \aistatsauthor{ Anonymous Author 1 \And Anonymous Author 2 \And Anonymous Author 3 }
% \aistatsaddress{ Unknown Institution 1 \And Unknown Institution 2 \And Unknown Institution 3 } 

\aistatsauthor{Jason K. Johnson \And Praneeth Netrapalli \And Michael Chertkov}
\aistatsaddress{
Center for Nonlinear Studies \\
\& Theoretical Division T-4 \\
Los Alamos National Laboratory,\\
Los Alamos NM
\And
Department of Electrical and\\
Computer Engineering \\
The University of Texas at Austin,\\
Austin TX
\And 
Center for Nonlinear Studies \\
\& Theoretical Division T-4 \\
Los Alamos National Laboratory,\\
Los Alamos NM
}

% \aistatsauthor{
% \hspace{-2cm} Praneeth Netrapalli 
% \And 
% \hspace{1cm} Jason K. Johnson \;\; Michael Chertkov}
% \aistatsaddress{
% \hspace{-1.7cm} Department of Electrical and Computer Engineering \\
% \hspace{-2cm} The University of Texas at Austin, Austin TX
% \And 
% \hspace{.9cm} Center for Nonlinear Studies \& Theoretical Division T-4 \\
% \hspace{1cm} Los Alamos National Laboratory, Los Alamos NM}
]

\begin{abstract} 
Inference and learning of graphical models are both well-studied problems in
statistics and machine learning that have found many applications in science and
engineering. However, exact inference is intractable in general graphical
models, which suggests the problem of seeking the best approximation to a
collection of random variables within some tractable family of graphical models.
 In this paper, we focus our attention on the class of planar Ising models, for
which inference is tractable using techniques of statistical physics [Kac and
Ward; Kasteleyn]. Based on these techniques and recent methods for planarity
testing and planar embedding [Chrobak and Payne], we propose a simple greedy
algorithm for learning the best planar Ising model to approximate an arbitrary
collection of binary random variables (possibly from sample data). Given
the set of all pairwise correlations among variables, we select a planar graph and
optimal planar Ising model defined on this graph to best approximate that
set of correlations.  We demonstrate our method in some simulations and for
the application of modeling senate voting records.
\end{abstract}

\section{Introduction}\label{sec:intro}

Graphical models \cite{Lauritzen,MacKay} are widely used to represent the statistical relations among a
set of random variables. Nodes of the graph correspond to random variables and
edges of the graph represent statistical interactions among the variables. The
problems of inference and learning on graphical models are encountered in many
practical applications. The problem of inference is to deduce certain
statistical properties (such as marginal probabilities, modes etc.) of a given
set of random variables whose graphical model is known. It has wide applications
in areas such as error correcting codes, statistical physics and so on.
The problem of learning on the other hand is to deduce the graphical
model of a set of random variables given statistics (possibly from samples) of
the random variables. Learning is also a widely encountered problem in areas
such as biology, anthropology and so on. %references

A certain class of binary-variable graphical models with pairwise interactions
known as the \emph{Ising model} has been studied by physicists as a simple model
of order-disorder transitions in magnetic materials \cite{onsager}. 
Remarkably, it was found that in the special case of an Ising model with
zero-mean $\{-1,+1\}$ binary random variables and pairwise interactions defined
on a planar graph, calculation of the partition function (which is closely tied
to inference) is tractable, essentially reducing to calculation of a matrix
determinant (\cite{kacward,sherman,kasteleyn,fisher}).
These methods have recently found uses in machine learning \cite{schraudolph,globerson}.

In this paper, we address the problem of approximating a collection of binary
random variables (given their pairwise marginal distributions) by a zero-mean
planar Ising model.  We also consider the related problem of selecting a
non-zero mean Ising model defined on an outer-planar graph (these models are
also tractable, being essentially equivalent to a zero-field model on a related
planar graph).

There has been a great deal of work on learning graphical models. Much of these
has focused on learning over the class of thin graphical models
\cite{bach,karger,shahaf} for which inference is tractable by converting the
model to tree-structured model.  The simplest case of this is learning tree
models (treewidth one graphs) for which it is tractable to find the best tree
model by reduction to a max-weight spanning tree problem \cite{chowliu}. 
However, the problem of finding the best bounded-treewidth model is NP-hard for
treewidths greater than two \cite{karger}, and so heuristic methods are used to
select the graph structure.  One popular method is to use convex optimization of
the log-likelihood penalized by $\ell_1$ norm of parameters of the graphical
model so as to promote sparsity \cite{banerjee,lee}. To go beyond low treewidth
graphs, such methods either focus on Gaussian graphical models or adopt a
tractable approximation of the likelihood.  Other methods seek only to learn the
graph structure itself \cite{ravikumar,abbeel} and are often able to demonstrate
asymptotic correctness of this estimate under appropriate conditions.  One
useful application of learning Ising models is for modeling interactions among
neurons \cite{cocco}.

The rest of the paper is organized as follows: We present the requisite
mathematical preliminaries in Section 2. Section 3 contains our algorithm along
with estimates of its computational complexity. We present simulation results in
Section 4 and an application to the senate voting record in Section 5.  We conclude
in Section 5 and suggest promising directions for further
research and development. All the proofs of propositions are delegated to an
appendix.  %to the extended version of the paper (provided as supplementary
%material to the reviewers; to be made available online upon publication).

\section{Preliminaries}\label{sec:prelim}

In this section, we develop our notation and briefly review the necessary
background theory. We will be dealing with binary random variables throughout
the paper. We write $P(x)$ to denote the probability distribution of a
collection of random variables $x=(x_1,\dots,x_n)$. Unless otherwise stated, we
work with undirected graphs $G=(V,E)$ with vertex (or node) set $V$ and edges
$\{i,j\} \in E \subset {V \choose 2}$. For vertices $i,j \in V$ we write $G+ij$
to denote the graph $(V,E \cup \{i,j\})$.

A \emph{(pairwise) graphical model} is a probability distribution $P(x) =
P(x_1,\dots,x_n)$ that is defined on a graph $G = (V,E)$ with vertices $V =
\{1,..,n\}$ as
\begin{eqnarray} \label{eq:graphical_model}
P(x) &\propto& \prod_{i \in V} \psi_i(x_i) \prod_{\{i,j\} \in E} \psi_{ij}(x_i,x_j) \nonumber \\
&\propto& \exp\{ \sum_{i \in V} f_i(x_i) + \sum_{\{i,j\} \in E} f_{ij}(x_i,x_j) \}
\end{eqnarray}
where $\psi_i,\psi_{ij} \geq 0$ are non-negative node and edge compatibility
functions. For positive $\psi$'s, we may also represent $P(x)$ as a Gibbs
distribution with potentials $f_i = \log \psi_i$ and $f_{ij} = \log \psi_{ij}$.

\subsection{Entropy, Divergence and Likelihood}\label{subsec:entropy}

For any probability distribution $P$ on some sample space $\chi$, its 
\emph{entropy} is defined as \cite{cover}
\begin{equation*}
 H(P) = -\displaystyle \sum_{x\in \chi} P(x) \log P(x)
\end{equation*}
Suppose we want to calculate how well a probability distribution $Q$
approximates another probability distribution $P$ (on the same sample space
$\chi$). For any two probability distributions $P$ and $Q$ on some sample space
$\chi$, we denote by $D(P,Q)$ the \emph{Kullback-Leibler divergence} (or
\emph{relative entropy}) between $P$ and $Q$.
\begin{equation*}\nonumber
D(P,Q) = \displaystyle \sum_{x \in \chi} P(x) \log \frac{P(x)}{Q(x)}
\end{equation*}
The \emph{log-likelihood function} is defined as follows:
\begin{equation*}\nonumber
LL(P,Q) = \displaystyle \sum_{x \in \chi} P(x) \log Q(x)
\end{equation*}
The probability distribution in a family $\mathcal{F}$ that maximizes the
log-likelihood of a probability distribution $P$ is called the
\emph{maximum-likelihood estimate} of $P$ in $\mathcal{F}$, and this is
equivalent to the \emph{minimum-divergence projection} of $P$ to $\mathcal{F}$:
\begin{equation}\nonumber
P_\mathcal{F} 
= \displaystyle \operatorname*{arg\,max}_{Q \in \mathcal{F}} LL(P,Q) 
= \displaystyle \operatorname*{arg\,min}_{Q \in \mathcal{F}} D(P,Q)
\end{equation}

\subsection{Exponential Families}\label{subsec:expfam}

A set of random variables $x = (x_1,\dots,x_n)$ are said to belong to an
\emph{exponential family} \cite{barndorff,wainwright} if there exist functions $\phi_1,\cdots, \phi_m$ (the
\emph{features} of the family) and scalars (\emph{parameters}) $
\theta_1,\cdots,\theta_m$ such that the joint probability distribution on the
variables is given by
\begin{equation*}
 P(x) = \frac{1}{Z(\theta)} \exp\left(\displaystyle \sum_\alpha \theta_\alpha \phi_\alpha(x) \right)
\end{equation*}
where $Z(\theta)$ is a normalizing constant called the \emph{partition
function}. This corresponds to a graphical model if $\phi_\alpha$ happen to be
functions on small subsets (e.g., pairs) of all the $n$ variables. The graph
corresponding to such a probability distribution is the graph where two nodes
have an edge between them if and only if there exists $\alpha$ such that
$\phi_\alpha$ depends on both variables. If the functions $\{\phi_\alpha\}$ are
non-degenerate (please refer to \cite{wainwright} for details), then for any achievable
moment parameters $\mu = \mathbb{E}_P[\phi]$ (for an arbitrary distribution $P$)
there exists a unique parameter vector $\theta(\mu)$ that realizes these moments
within the exponential family.

Let $\Phi(\theta)$ denote the \emph{log-partition function}
\begin{equation*}
 \Phi(\theta) \triangleq \log Z(\theta)
\end{equation*}
For an exponential family, we have the following important relation (of
\emph{conjugate duality}) between the log-partition function and negative entropy of the
corresponding probability distribution $H(\mu) \triangleq H(P_{\theta(\mu)})$ as
follows \cite{wainwright}
\begin{equation}\label{eq:ml_convex_opt}
 \Phi^*(\mu) \triangleq \displaystyle \max_{\theta \in \mathbb{R}^m} \left\{ \mu^T \theta - \Phi(\theta) \right\} = -H(\mu)
\end{equation}
if the mean parameters $\mu$ are achievable under some probability distribution.
In fact, this corresponds to the problem of maximizing the log-likelihood
relative to an arbitrary distribution $P$ with moments $\mu =
\mathbb{E}_P[\phi]$ over the exponential family. The optimal choice of $\theta$
realizes the given moments ($\mathbb{E}_\theta[\phi] = \mu$) and this solution
is unique for non-degenerate choice of features. 

%It is also known that both
%$\Phi(\theta)$ and $\Phi^*(\mu)$ are convex functions. Hence, if it is tractable
%to compute $\Phi(\theta)$ of the family then it is likewise tractable to compute
%$H(\mu)$ by solving a tractable convex optimization problem. 

\subsection{Ising Graphical Model}\label{subsec:brush}

The Ising model is a famous model in statistical physics that
has been used as simple model of magnetic phenomena and of phase transitions in
complex systems.
\begin{defn}
An \emph{Ising model} on binary random variables $x = (x_1,\dots,x_n)$ and graph
$G=(V,E)$ is the probability distribution defined by 
\begin{displaymath}
P(x) = \frac{1}{Z(\theta)} \exp\left(\sum_{i \in V} \theta_i x_i +\sum_{\{i,j\} \in E}\theta_{ij}x_ix_j\right)
\end{displaymath}
where $x_i\in \{-1,1\}$. Thus, the model is specified by vertex parameters
$\theta_i$ and edge parameter $\theta_{ij}$.
\end{defn}

This defines an exponential family with non-degenerate features $(\phi_i(x)=x_i,
i \in V)$ and $(\phi_{ij}(x) = x_i x_j, \{i,j\} \in E)$ and with corresponding
moments $(\mu_i = \mathbb{E}[x_i], i \in V)$ and $(\mu_{ij} = \mathbb{E}[x_i
x_j], \{i,j\} \in E)$.

In fact, any graphical model with binary variables and soft pairwise potentials
can be represented as an Ising model with binary variables $x_i = \{-1,+1\}$ and
with parameters
\begin{displaymath}
\theta_i = \tfrac{1}{2} \sum_{x_i} x_i f_i(x_i) 
+ \tfrac{1}{4} \sum_{\{i,j\} \in E} \sum_{x_i,x_j} x_i f_{ij}(x_i,x_j) \\
\end{displaymath}
\begin{displaymath}
\theta_{ij} = \tfrac{1}{4} \sum_{x_i,x_j} x_i x_j f_{ij}(x_i,x_j). 
\end{displaymath}
There is also a simple correspondence between the moment parameters of the Ising
model and the node and edge-wise marginal distributions. Of course, it is
trivial to compute the moments given these marginals: $\mu_i = \sum_{x_i} x_i
P(x_i)$ and $\mu_{ij} = \sum_{x_i,x_j} x_i x_j P(x_i,x_j)$. The marginals are
recovered from the moments by:
\begin{displaymath}
P(x_i) = \tfrac{1}{2} (1 + \mu_i x_i)
\end{displaymath}
\begin{displaymath}
P(x_i,x_j) = \tfrac{1}{4} (1 + \mu_i x_i + \mu_j x_j + \mu_{ij} x_i x_j)
\end{displaymath}

We will be especially concerned with the following sub-family of Ising models:
\begin{defn}
An Ising model is said to be \emph{zero-field} if $\theta_i = 0$ for all $i \in
V$. It is \emph{zero-mean} if $\mu_i = 0$ ($P(x_i = \pm 1) = \tfrac{1}{2}$) for
all $i \in V$.
\end{defn}
It is simple to verify that the Ising model is zero-field if and only if it is
zero-mean. Although the assumption of zero-field appears very restrictive, a
general Ising model can be represented as a zero-field model by adding one
auxiliary variable node connected to every other node of the graph \cite{globerson}. The
parameters and moments of the two models are then related as follows:
\begin{prop}\label{prop:zeromeantononzero}
Consider the Ising model on $G=(V,E)$ with $V = \{1,\dots,n\}$, parameters
$\{\theta_i\}$ and $\{\theta_{ij}\}$, moments $\{\mu_i\}$ and $\{\mu_{ij}\}$ and
partition function $Z$. Let $\widehat{G} = (\widehat{V},\widehat{E})$ denote the extended
graph based on nodes $\widehat{V} = V \cup \{n+1\}$ with edges $\widehat{E} = E \cup
\{\{i,n+1\}, i \in V \})$. We define a zero-field Ising model on $\widehat{G}$ with
parameters $\{\widehat\theta_{ij}\}$, moments $\{\widehat\mu_{ij}\}$ and partition
function $\widehat{Z}$. If we set the parameters according to
\begin{displaymath}
\widehat\theta_{ij} =
\left\{
\begin{array}{ll}
 \theta_i ~\mbox{if}~ j=n+1\\
 \theta_{ij} ~\mbox{otherwise}
\end{array}
\right.
\end{displaymath}
then $\widehat{Z} = 2 Z$ and 
\begin{displaymath}
\widehat\mu_{ij} =
\left\{
\begin{array}{ll}
 \mu_i ~\mbox{if}~ j=n+1\\
 \mu_{ij} ~\mbox{otherwise}
\end{array}
\right.
\end{displaymath}
\end{prop}
Thus, inference on the corresponding zero-field Ising model on the extended
graph $\widehat{G}$ is essentially equivalent to inference on the (non-zero-field)
Ising model defined on $G$.

\subsection{Inference for Planar Ising Models}\label{subsec:partfunc_planarising}

The motivation for our paper is the following result on tractability of inference 
for the \emph{planar zero-field Ising model}.
\begin{defn} 
A graph is \emph{planar} if it may be embedded in the plane without any edge crossings. 
\end{defn}
Moreover, it is known that any planar graph can be embedded such that all edges 
are drawn as straight lines.

\begin{theorem}\label{thm:partfunc_planarising}
\cite{kacward}\cite{sherman}
Let $Z$ denote the partition function of a zero-field Ising model defined on a
planar graph $G=(V,E)$. Let $G$ be embedded in the plane (with edges drawn as
straight lines) and let $\phi_{ijk} \in [-\pi,+\pi]$ denote the angular
difference (turning angle) between directed edges $(i,j)$ and $(j,k)$. We define
the matrix $W \in \mathbb{C}^{2 |E| \times 2 |E|}$ indexed by directed edges of
the graph as follows: $W = A D$ where $D$ is the diagonal matrix with $D_{ij,ij}
= \tanh \theta_{ij} \triangleq w_{ij}$ and
\begin{displaymath}
A_{ij,kl} = 
\left\{ 
\begin{array}{ll}
\exp(\tfrac{1}{2} \sqrt{-1} \phi_{ijl}), &  j=k ~\mbox{and}~ i \neq l \\
0, & \mbox{otherwise}
\end{array}
\right.
\end{displaymath}
Then, the partition function of the zero-field planar Ising model is given by:
\begin{displaymath}
 Z = 2^n \left(\prod_{\{i,j\}\in E} \cosh \theta_{ij}\right) \det(I-W)^{\frac{1}{2}}
\end{displaymath}
\end{theorem}

We briefly remark the combinatorial interpretation of this theorem: $W$ is the
generating matrix of non-reversing walks of the graph with the weight of a walk
$\gamma$ being 
\begin{displaymath}
w_\gamma = \prod_{(i,j) \in \gamma} w_{ij} \prod_{(i,j,k) \in \gamma} \exp(\tfrac{1}{2} \sqrt{-1} \phi_{ijk}).
\end{displaymath}
The determinant can be interpreted as the (inverse) graph zeta function: 
$\det(I-W) = \prod_\gamma (1-w_\gamma)$
where the product is taken over all equivalence classes of aperiodic closed
non-reversing walks \cite{sherman,loebl}. A related method for computing the Ising model partition function is 
based on counting perfect matching of planar graphs \cite{kasteleyn,fisher}.  
We favor the Kac-Ward approach only because it is somewhat more direct. 

Since calculation of the partition function reduces to calculating the
determinant of a matrix, one may use standard Gaussian elimination methods to
evaluate this determinant with complexity $O(n^3)$. In fact, using the
generalized nested dissection algorithm to exploit sparsity of the matrix, the
complexity of these calculations can be reduced to $O(n^{3/2})$ \cite{liptonrose,liptontarjan,galluccio}.
Thus, inference of the zero-field planar Ising model is tractable and scales well with
problem size.

It also turns out that the gradient and Hessian of the log-partition function
$\Phi(\theta) = \log Z(\theta)$ can be calculated efficiently from the Kac-Ward
determinant formula. We recall that derivatives of $\Phi(\theta)$ recover the
moment parameters of the exponential family model \cite{barndorff,wainwright}: 
\begin{displaymath}
\nabla \Phi(\theta) = \mathbb{E}_\theta[\phi] = \mu.
\end{displaymath}
Thus, inference of moments (and node and edge marginals) are likewise tractable
for the zero-field planar Ising model.

\begin{prop}\label{prop:gradienthessiancalc} 
Let $\mu = \nabla \Phi(\theta)$, $H = \nabla^2 \Phi(\theta)$. Let $S =
(I-W)^{-1} A$ and $T = (I+P) (S \circ S^T) (I+P^T)$ where $A$ and $W$ are defined as in
Theorem 1, $\circ$ denotes the element-wise product and $P$ is the permutation matrix
swapping indices of directed edges $(i,j)$ and $(j,i)$. Then,
\begin{displaymath}
\mu_{ij} =  w_{ij} - \tfrac{1}{2} (1-w^2_{ij}) (S_{ij,ij} + S_{ji,ji}) 
\end{displaymath}
\begin{displaymath}
H_{ij,kl} = 
\left\{
\begin{array}{ll}
1-\mu_{ij}^2 ~\mbox{if}~ ij = kl, ~\mbox{else} \\
- \tfrac{1}{2} (1-w^2_{ij}) T_{ij,kl} (1-w^2_{kl})
\end{array}
\right.
\end{displaymath}
\end{prop}
 
Note, calculating the full matrix $S$ requires $O(n^3)$ calculations. However,
to compute just the moments $\mu$ only the diagonal elements of $S$ are needed. 
Then, using the generalized nested dissection method, inference of moments
(edge-wise marginals) of the zero-field Ising model can be achieved with
complexity $O(n^{3/2})$.
However, computing the full Hessian is more expensive, requiring $O(n^3)$ calculations.

\paragraph{Inference for Outer-Planar Graphical Models}

We emphasize that the above calculations require both a planar graph $G$ and a
zero-field Ising model. Using the graphical transformation of Proposition 1,
the latter zero-field condition may be relaxed but at the expense of adding an
auxiliary node connected to all the other nodes. In general planar graphs $G$,
the new graph $\widehat{G}$ may not be planar and hence may not admit tractable
inference calculations. However, for the subset of planar graphs where this
transformation does preserve planarity inference is still tractable.

\begin{defn} A graph $G$ is said to be \emph{outer-planar} if there exists an
embedding of $G$ in the plane where all the nodes are on the outer face.
\end{defn}

In other words, the graph $G$ is outer-planar if the extended graph $\widehat{G}$
(defined by Proposition 1) is planar. Then, from Proposition 1 and Theorem 1 it
follows that:

\begin{prop} \cite{globerson} The partition function and moments of any outer-planar Ising
graphical model (not necessarily zero-field) can be calculated efficiently. 
Hence, inference is tractable for any binary-variable graphical model with
pairwise interactions defined on an outer-planar graph. \end{prop}

This motivates the problem of learning outer-planar graphical models
for a collection of (possibly non-zero mean) binary random variables.

\section{Learning Planar Ising Models}\label{sec:main}

This section addresses the main goals of the paper, which are two-fold:
\begin{enumerate}
\item Solving for the maximum-likelihood Ising model on a given planar graph
to best approximate a collection of zero-mean random variables.
\item How to select (heuristically) the planar graph to obtain the best
approximation.
\end{enumerate}
We address these respective problems in the following 
two subsections. The solution of the first problem is an integral
part of our approach to the second. Both solutions are easily
adapted to the context of learning outer-planar graphical models of
(possibly non-zero mean) binary random variables.

\subsection{ML Parameter Estimation}\label{subsec:convexopt}

As discussed in Section 2.2, maximum-likelihood estimation over an exponential
family is a convex optimization problem (\ref{eq:ml_convex_opt}) based on the
log-partition function $\Phi(\theta)$. In the case of the zero-field Ising model
defined on a given planar graph it is tractable to compute $\Phi(\theta)$ via a
matrix determinant described in Theorem 1. Thus, we obtain an unconstrained,
tractable, convex optimization problem for the maximum-likelihood zero-field
Ising model on the planar graph $G$ to best approximate a probability
distribution $P(x)$:
\begin{displaymath}
\max_{\theta \in \mathbb{R}^{|E|}}
\{ \sum_{ij} (\mu_{ij} \theta_{ij} - \log\cosh \theta_{ij}) - \tfrac{1}{2} \log\det (I-W(\theta)) \}
\end{displaymath}
Here, $\mu_{ij} = \mathbb{E}_P[x_i x_j]$ for all edges $\{i,j\} \in G$ and the
matrix $W(\theta)$ is as defined in Theorem 1. If $P$ represents the empirical
distribution of a set of independent identically-distributed (iid) samples
$\{x^{(s)}, s=1,\dots,S\}$ then $\{\mu_{ij}\}$ are the corresponding empirical
moments $\mu_{ij} = \frac{1}{S} \sum_s x^{(s)}_i x^{(s)}_j$.

\paragraph{Newton's Method} 

We solve this unconstrained convex optimization problem using Newton's method
with step-size chosen by back-tracking line search \cite{boyd}. This produces a
sequence of estimates $\theta^{(s)}$ calculated as follows:
\begin{displaymath}
\theta^{(s+1)} = \theta^{(s)} + \lambda_s H(\theta^{(s)})^{-1} ( \mu(\theta^{(s)}) - \mu)
\end{displaymath}
where $\mu(\theta^{(s)})$ and $H(\theta^{(s)})$ are calculated using Proposition 2 and
$\lambda_s \in (0,1]$ is a step-size parameter chosen by backtracking line
search (see \cite{boyd} for details). The per iteration complexity of this
optimization is $O(n^3)$ using explicit computation of the Hessian at each
iteration. This complexity can be offset somewhat by only re-computing the
Hessian a few times (reusing the same Hessian for a number of iterations), to
take advantage of the fact that the gradient computation only requires
$O(n^\frac{3}{2})$ calculations. As Newton's method has quadratic
convergence, the number of iterations required to achieve a high-accuracy
solution is typically 8-16 iterations (essentially independent of problem size).
 We estimate the computational complexity of solving this convex optimization
problem as roughly $O(n^3)$.

\subsection{Greedy Planar Graph Selection}\label{subsec:problem}

We now consider the problem of selection of the planar graph $G$ to best
approximate a probability distribution $P(x)$ with pairwise moments $\mu_{ij} =
\mathbb{E}_P[x_i x_j]$ given for all $i,j \in V$. Formally, we seek the planar
graph that maximizes the log-likelihood (minimizes the divergence) relative to
$P$:
\begin{displaymath}
\widehat{G} 
= \operatorname*{arg\,max}_{G \in \mathcal{P}_V} LL(P,P_G) 
= \operatorname*{arg\,max}_{G \in \mathcal{P}_V} \max_{Q \in \mathcal{F}_G} LL(P,Q)
\end{displaymath}
where $\mathcal{P}_V$ is the set of planar graphs on the vertex set $V$,
$\mathcal{F}_G$ denotes the family of zero-field Ising models defined on graph
$G$ and $P_G = \operatorname*{arg\,max}_{Q \in \mathcal{F}_G} LL(P,Q)$ is the maximum-likelihood
(minimum-divergence) approximation to $P$ over this family.

We obtain a heuristic solution to this graph selection problem using the
following greedy edge-selection procedure. The input to the algorithm is a
probability distribution $P$ (which could be empirical) on $n$ binary $\{-1,1\}$
random variables. In fact, it is sufficient to summarize $P$ by its pairwise
correlations $\mu_{ij} = \mathbb{E}_P[x_i x_j]$ on all pairs $i,j \in V$. The
output is a maximal planar graph $G$ and the maximum-likelihood approximation
$\theta_G$ to $P$ in the family of zero-field Ising models defined on this
graph.

\begin{algorithm}[ht]
\caption{GreedyPlanarGraphSelect($P$)}
\label{GLPI}
\begin{algorithmic}[1]
 \STATE $G = \emptyset, \theta_G = 0$
 \FOR {$k = 1:3n-6$} 
   \STATE $\Delta = \left\{ \{i,j\} \subset V |  \{i,j\} \notin G, G+ij \in \mathcal{P}_V \right\}$
   \STATE $\tilde\mu_\Delta = \{ \tilde\mu_{ij} = \mathbb{E}_{\theta_G}[x_i x_j], \{i,j\} \in \Delta \}$ 
   \STATE $G \leftarrow G \cup \displaystyle \operatorname*{arg\,max}_{e \in \Delta} D(P_e, \tilde{P}_e)$
   \STATE $\theta_G = \mbox{PlanarIsing}(G,P$)
\ENDFOR
\end{algorithmic}
\end{algorithm}

The algorithm starts with an empty graph and then sequentially adds edges to the
graph one at a time so as to (heuristically) increase the log-likelihood
(decrease the divergence) relative to $P$ as much as possible at each step. 
Here is a more detailed description of the algorithm along with estimates of the
computational complexity of each step: 
\begin{itemize}

\item \emph{Line 3.} First, we enumerate the set $\Delta$ of all edges one
might add (individually) to the graph while preserving planarity. This is
accomplished by an $O(n^3)$ algorithm in which we iterate over all pairs
$\{i,j\} \not\in G$ and for each such pair we form the graph $G + ij$ and
test planarity of this graph using known $O(n)$ algorithms \cite{chrobak}.

\item \emph{Line 4.} Next, we perform tractable inference calculations with
respect to the Ising model on $G$ to calculate the pairwise correlations
$\tilde{\mu}_{ij}$ for all pairs $\{i,j\} \in \Delta$. This is accomplished
using $O(n^{3/2})$ inference calculations on augmented versions of the graph
$G$. For each inference calculation we add as many edges to $G$ from $\Delta$
as possible (setting $\theta = 0$ on these edges) while preserving planarity and
then calculate all the edge-wise moments of this graph using Proposition 2
(including the zero-edges). This requires at most $O(n)$ iterations to cover
all pairs of $\Delta$, so the worst-case complexity to compute all required
pairwise moments is $O(n^{5/2})$.

\item \emph{Line 5.} Once we have these moments, which specify the corresponding
pairwise marginals of the current Ising model, we compare these moments
(pairwise marginals) to those of the input distribution $P$ by evaluating the
pairwise KL-divergence between the Ising model and $P$. As seen by the
following proposition, this gives us a lower-bound on the improvement obtained
by adding the edge:
\begin{prop}\label{prop:lboundKLD}
Let $P_G$ and $P_{G+ij}$ be the projections of $P$ on $G$ and $G+ij$ respectively. 
Then,
\begin{displaymath}
D(P, P_{G}) - D(P, P_{G+ij}) \geq D\left( P(x_i,x_j) , P_G(x_i,x_j) \right) 
\end{displaymath}
where $P(x_i,x_j)$ and $P_G(x_i,x_j)$ represent the marginal distributions on $x_i,x_j$ of 
probabilities $P$ and $P_G$ respectively.
\end{prop}
Thus, we greedily select the next edge $\{i,j\}$ to add so as to maximize this
lower-bound on the improvement measured by the increase on log-likelihood (this
being equal to the decrease in KL-divergence).

\item \emph{Line 6.} Finally, we calculate the new maximum-likelihood parameters
$\theta_G$ on the new graph $G \leftarrow G +ij$. This involves solution
of the convex optimization problem discussed in the preceding subsection, which
requires $O(n^3)$ complexity. This step is necessary in order to subsequently
calculate the pairwise moments $\tilde\mu$ which guide further edge-selection
steps, and also to provide the final estimate.

\end{itemize}

We continue adding one edge at a time until a maximal planar graph (with $3n-6$
edges) is obtained. Thus, the total complexity of our greedy algorithm for
planar graph selection is $O(n^4)$.

\paragraph{Non-Maximal Planar Graphs}

Since adding an edge always gives an improvement in the log-likelihood, the
greedy algorithm always outputs a maximal planar graph. However, this might
lead to over-fitting of the data
%This might not be desirable
especially when the input probability distribution corresponds to
an empirical distribution.
%Since finite samples always lead to some estimation
% error, the maximal planar graph may tend to overfit the data and therefore give
% a poor estimate to the true distribution being sampled.
In such cases, to avoid
over-fitting, we might modify the algorithm so that an
edge is added to the graph only if the improvement in log-likelihood is more
than some threshold $\gamma$.
% For a threshold value of $\gamma=0$, the algorithm
% outputs a maximal planar graph.  But for larger values of $\gamma$ the output
% graph will become sparser.
An experimental search can be performed for a
suitable value of this threshold (e.g. so as to minimize some estimate of the
generalization, such as in cross validation methods \cite{zhang}).
% or one
% could use some heuristic value for $\gamma$ based on the number of samples such
% as Akaike's information criterion (AIC) or Shwarz's Bayesian information
% criterion (BIC).%\cite{akaike,schwarz}.

\paragraph{Outer-Planar Graphs and Non-Zero Means}

The greedy algorithm returns a zero-field Ising model (which has zero mean for
all the random variables) defined on a planar graph. If the actual random
variables are non-zero mean, this may not be desirable. For this case we may
prefer to exactly model the means of each random variable but still retain
tractability by restricting the greedy learning algorithm to select outer-planar
graphs. This model faithfully represents the marginals of each random variable
but at the cost of modeling fewer pairwise interactions among the variables.  

This is equivalent to the following procedure. First, given the sample moments
$\{\mu_i\}$ and $\{\mu_{ij}\}$ we convert these to an equivalent set of
zero-mean moments $\widehat\mu$ on the extended vertex set $\widehat{V} = V \cup
\{n+1\}$ according to Proposition 1. Then, we select a zero-mean planar Ising
model for these moments using our greedy algorithm. However, to fit the means
of each of the original $n$ variables, we initialize this graph to include all
the edges $\{i,n+1\}$ for all $i \in V$.
%(requiring that these are present in our final estimate of the graph $\widehat{G}$).
After this initialization step, we use
the same greedy edge-selection procedure as before. This yields the graph
$\widehat{G}$ and parameters $\theta_{\widehat{G}}$. Lastly, we convert back to a
(non-zero field) Ising model on the subgraph of $\widehat{G}$ defined on nodes $V$,
as prescribed by Proposition 1. The resulting graph $G$ and parameters
$\theta_G$ is our heuristic solution for the maximum-likelihood outer-planar 
Ising model.  

Lastly, we remark that it is not essential that one chooses between the
zero-field planar Ising model and the outer-planar Ising model.  We may allow
the greedy algorithm to select something in between---a partial outer-planar
Ising model where only nodes of the outer-face are allowed to have non-zero
means. This is accomplished simply by omitting the initialization step of
adding edges $\{i,n+1\}$ for all $i \in V$.
% , that is, by initializing the greedy
% algorithm for learning $\widehat{G}$ with the disconnected graph on $\widehat{V}$.

\section{Simulations}

In this section, we present the results of numerical experiments evaluating our algorithm.

\paragraph{Counter Example}

The first result, presented in Figure $1$ illustrates the fact that our algorithm does not always recover the exact
structure even when the underlying graph is planar and the algorithm is given exact moments as
inputs.
%Figure $1$ presents the counter example.
% The basic idea is that graphical models can have nodes which are not neighbors but are more correlated
% than some other nodes which are neighbors. If the spurious edges corresponding to these highly correlated nodes
% are added early on in the algorithm, then the actual edges may have to be left out because of the planarity
% restriction.

\begin{figure}[ht]
\centering
\subfigure[]%[Original graphical model]
{
\includegraphics[width=3cm]{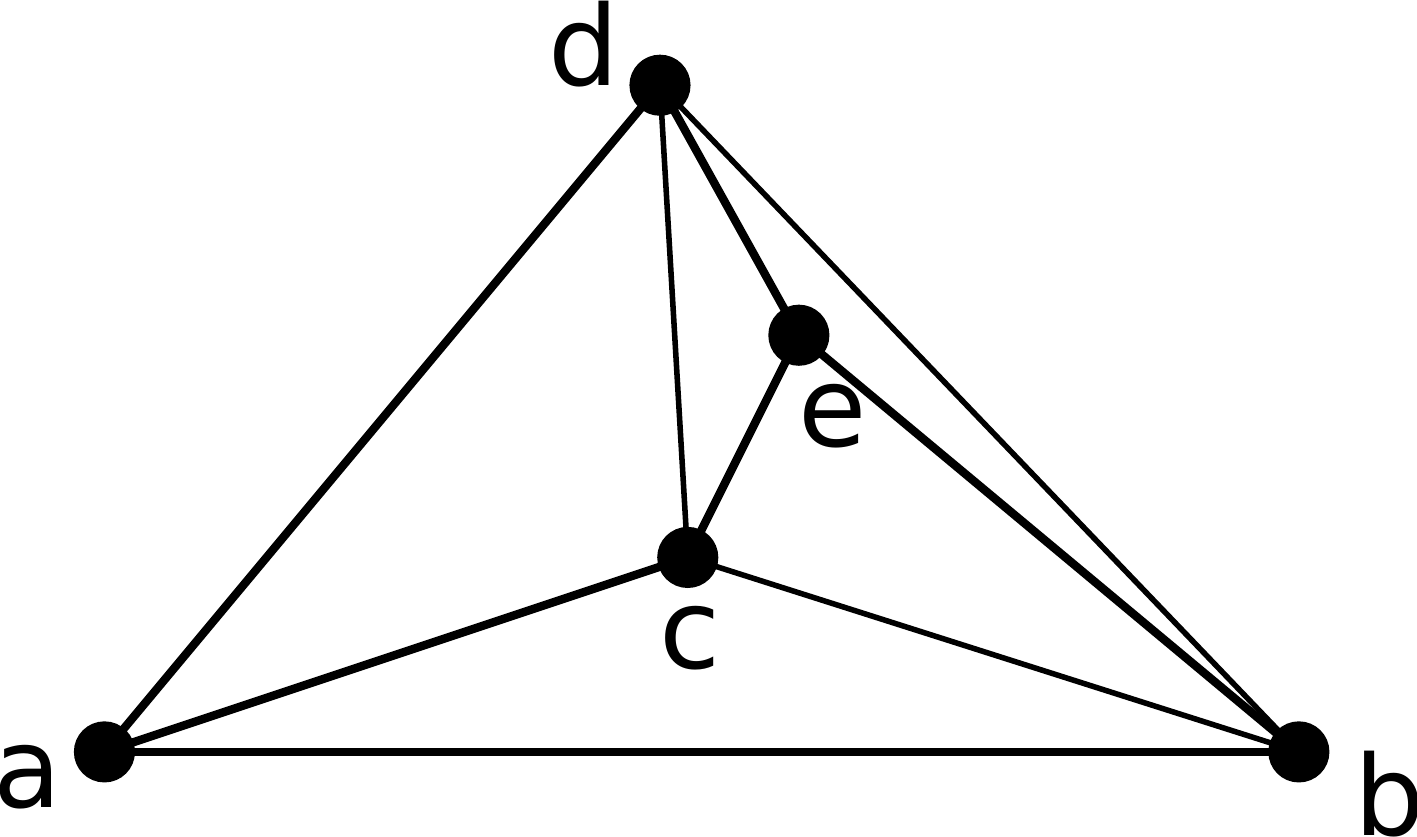}
\label{fig:counterexample_input}
}
% \subfigure[Correlation graph]{
% \includegraphics[width=3.5cm]{figures/counterexample_moments.jpg}
% \label{fig:counterexample_moments}
% }
\subfigure[]%[Recovered model]
{
\includegraphics[width=3.5cm]{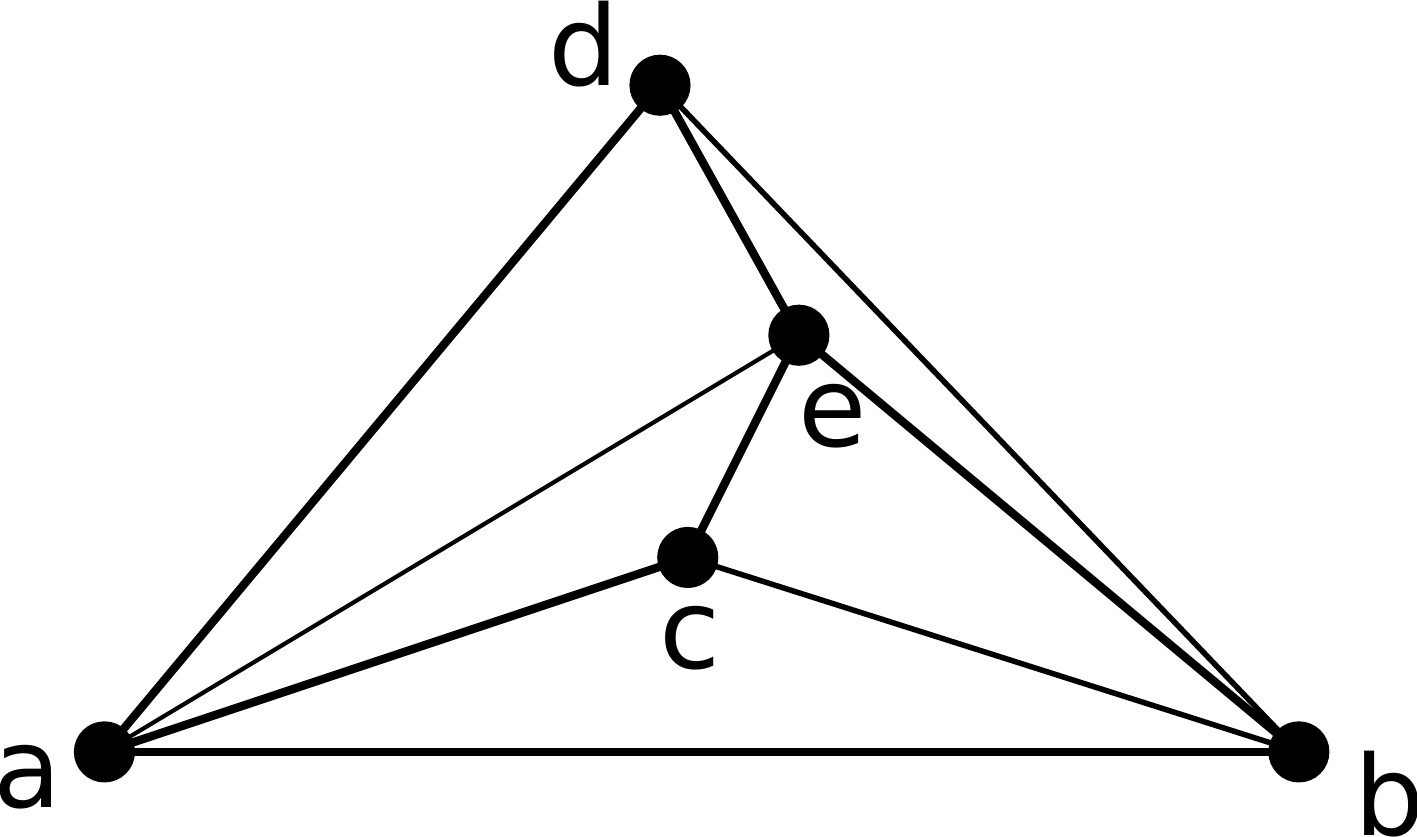}
\label{fig:counterexample_output}
}
\label{fig:counterexample}
\vspace{-.3cm}
\caption{Counter example : (a) Original graphical model (b) Recovered graphical model. The recovered graphical model
has one spurious edge $\{a,e\}$ and one missing edge $\{c,d\}$. It is clear from this example that our algorithm is not
always optimal.}
\end{figure}

We define a zero-field Ising model on the graph in Figure
\ref{fig:counterexample_input} with the edge parameters as follows:
$\theta_{bc}=\theta_{cd}=\theta_{bd}=0.1$ and $\theta_{ij}=1$ for all the other
edges. Figure \ref{fig:counterexample_input} shows the edge parameters in the
graph pictorially using the intensity of the edges - higher the intensity of an
edge, higher the corresponding edge parameter. When the edge parameters are as
chosen above, the correlation between nodes $a$ and $e$ is greater than the
correlation between any other pair of nodes. This leads to the edge between $a$
and $e$ to be the first edge added in the algorithm. However, since $K5$ (the
complete graph on $5$ nodes) is not planar, one of the actual edges
% (in this case, the edge between nodes $c$ and $d$) 
is missed in the output graph. Figure \ref{fig:counterexample_output}
shows the edge weighted recovered graph.% and again as before, the magnitude of the edge parameters is shown using the intensity of edges.
% It can be seen that once all the other edges are added as well, the edge parameter corresponding to the edge $\{a,e\}$
% is very small and hence if we were to consider deleting edges along with adding edges, we could still have gotten the
% original graph exactly.

\paragraph{Recovery of Zero-Field Planar Ising Model}

We now present the results of our experiments on a zero field Ising model on a
$7 \times 7$ grid. The edge parameters are chosen to be uniformly random between
$-1$ and $1$ with the condition that the absolute value be greater than a
threshold (chosen to be $0.05$) so as to avoid edges with negligible
interactions. We use Gibbs sampling to obtain samples from this model and
calculate empirical moments from these samples which are then passed as input to
our algorithm.  The results are seen in Figure 2 (see caption for details).
\begin{figure}[h]
\centering
\subfigure[]{
\includegraphics[width=2cm]{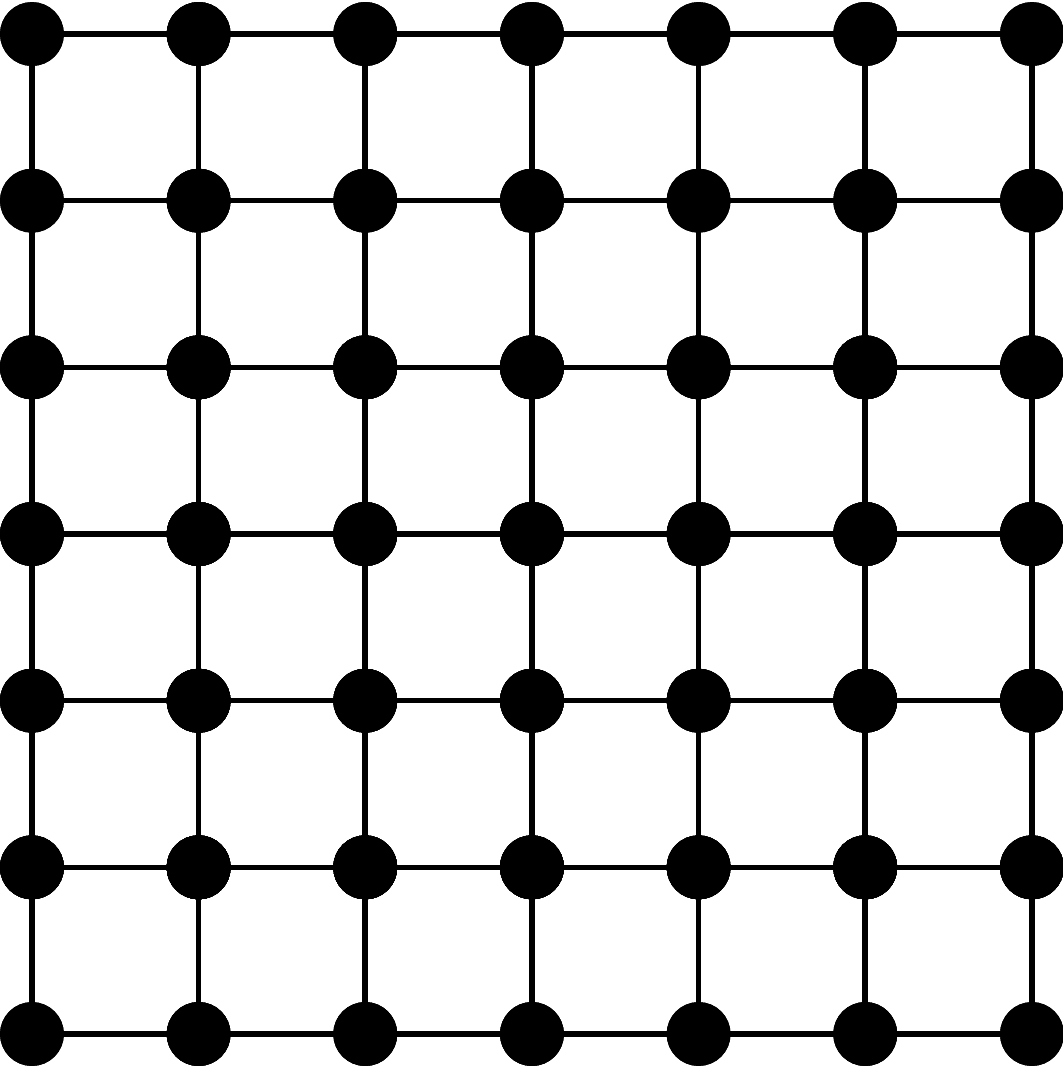}
\label{fig:7x7grid_input}
}
% \subfigure[]{
% \includegraphics[width=3cm,scale=.5]{figures/7x7grid_input_moments_goodtheta.jpg}
% \label{fig:7x7grid_moments}
% }
\subfigure[]{
\includegraphics[width=2cm]{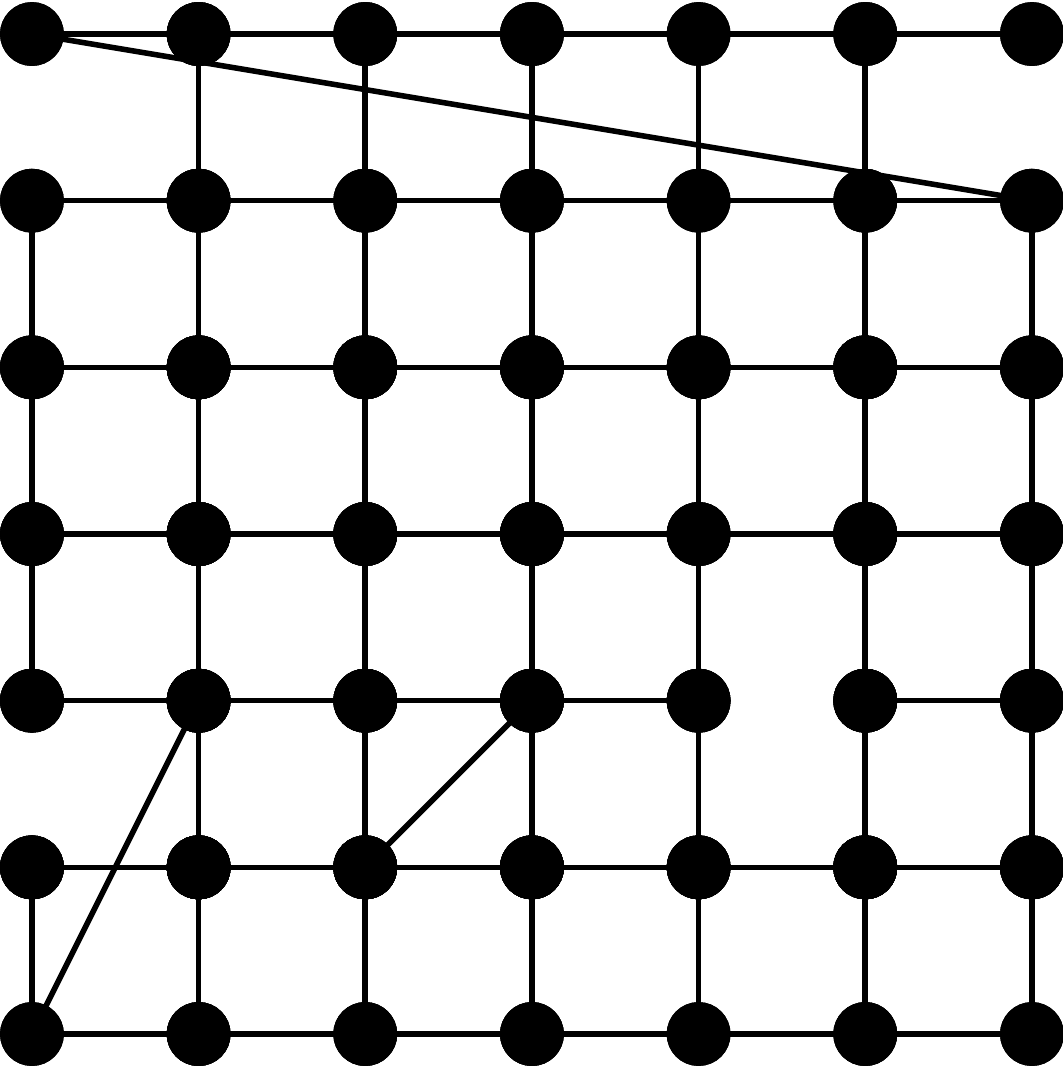}
\label{fig:7x7grid_output1}
}
\subfigure[]{
\includegraphics[width=2cm]{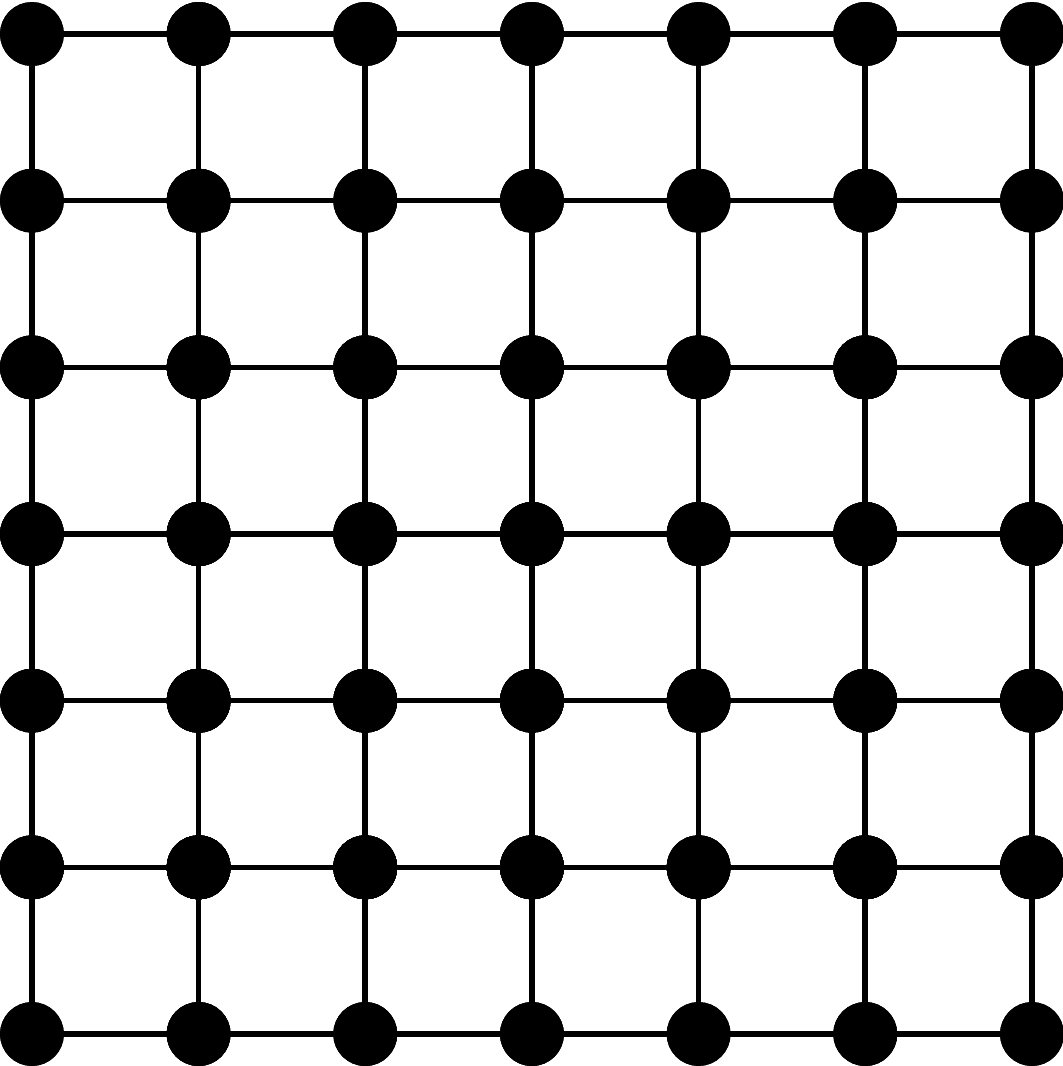}
\label{fig:7x7grid_output2}
}
\label{fig:7x7grid}
\vspace{-.3cm}
\caption{$7\times 7$ grid :
(a) Original graphical model
%(b) Input moments (calculated empirically from $10^5$ samples)
(b) Recovered graphical model ($10^4$ samples)
(c) Recovered graphical model ($10^5$ samples).
The inputs to the algorithm are the empirical moments obtained from the samples.
The algorithm is stopped when the recovered graph has the same number of edges as the original graphical model.
With $10^4$ samples, there are some errors in the recovered graphical model. When the number of samples
is increased to $10^5$, we see perfect recovery.
}
\end{figure}

\paragraph{Recovery of Non-Zero-Field Outer Planar Ising Model}

\begin{figure}[h]
\centering
\subfigure[]{
\includegraphics[width=2cm]{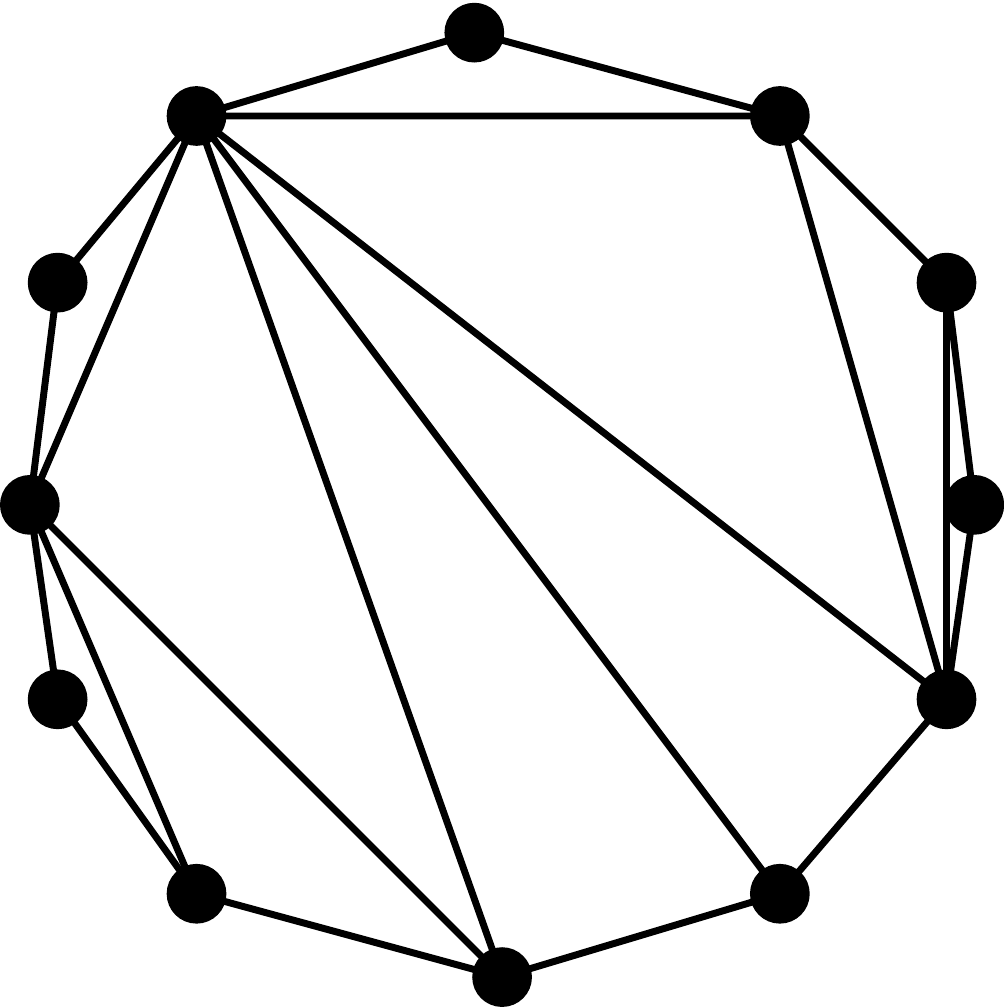}
\label{fig:outerplanar_input}
}
\subfigure[]{
\includegraphics[width=2cm]{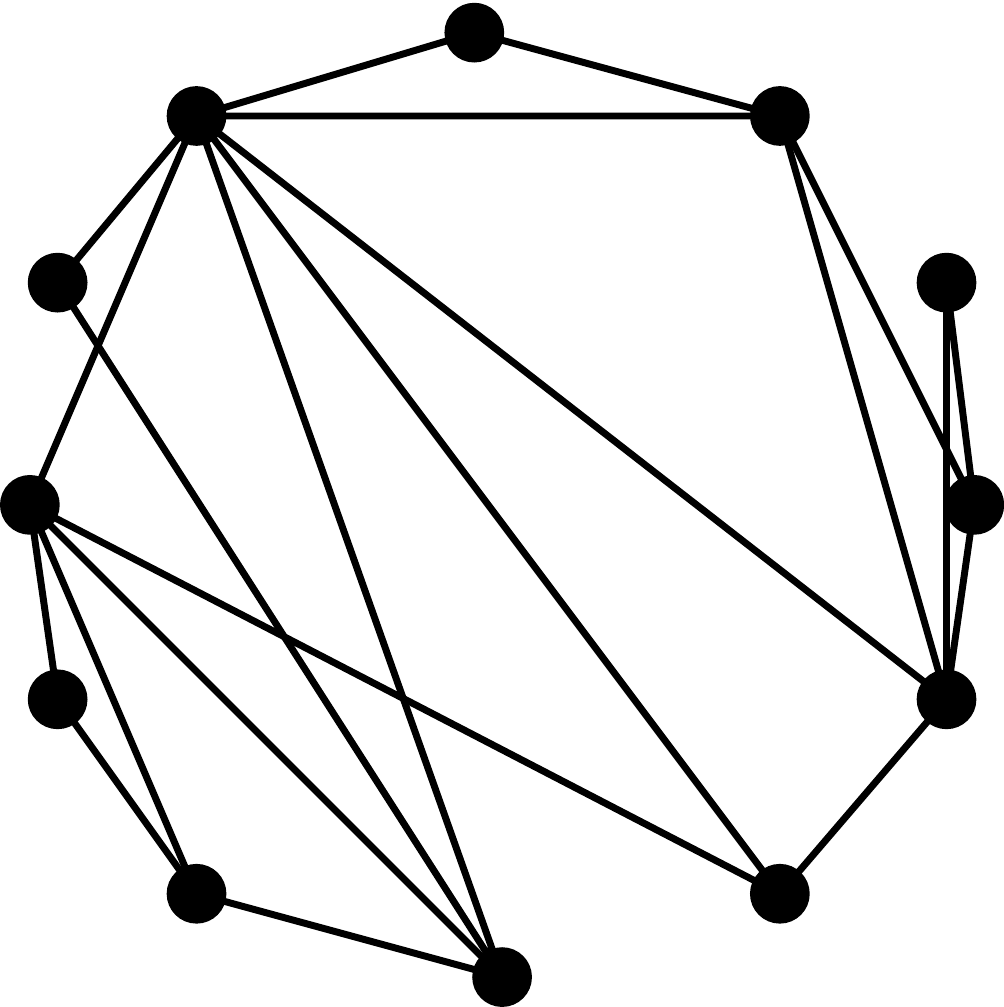}
\label{fig:outerplanar_output1}
}
\subfigure[]{
\includegraphics[width=2cm]{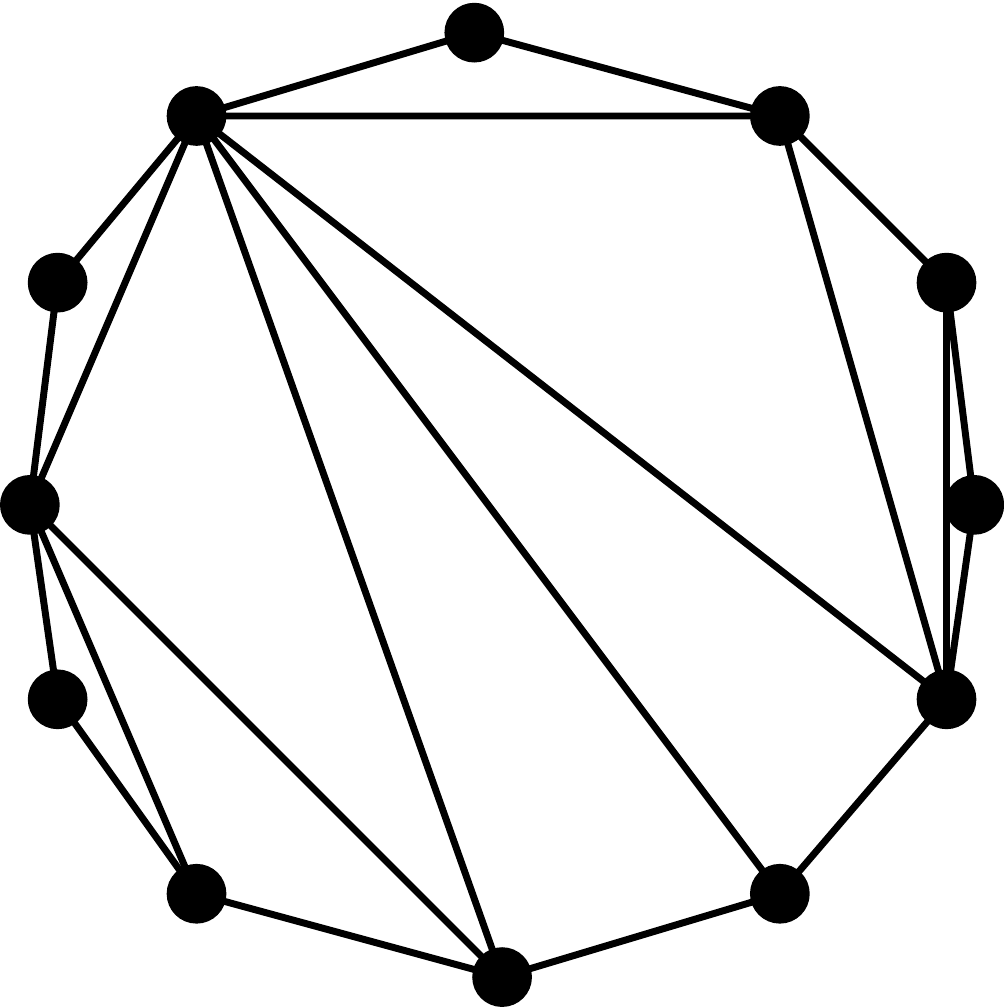}
\label{fig:outerplanar_output2}
}
\label{fig:outerplanar}
\vspace{-.3cm}
\caption{Outer planar graphical model :
(a) Original graphical model
(b) Recovered graphical model ($10^3$ samples)
(c) Recovered graphical model ($10^4$ samples).
Even in this case, the number of erroneous edges in the output of our algorithm decreases with increasing number of samples.
With $10^4$ samples, we recover the graphical model exactly.
}
\vspace{-.3cm}
\end{figure}

As explained in Section \ref{subsec:problem}, our algorithm can also be used to
find the best outer planar graphical model describing a given empirical
probability distribution. In this section, we present the results of our
numerical experiments on a $12$ node outer planar binary pairwise graphical
model where the nodes have non-zero mean. Though our algorithm gives perfect
reconstruction on graphs with many more nodes, we choose a small example to
illustrate the result effectively. We again use Gibbs sampling to obtain samples
and calculate empirical moments from those samples. Figure
\ref{fig:outerplanar_input} presents the original graphical model. Figures
\ref{fig:outerplanar_output1} and \ref{fig:outerplanar_output2} present the
output graphical models for $10^3$ and $10^4$ samples respectively. We make sure
that the first moments of all the nodes are satisfied by starting with the
auxiliary node connected to all other nodes. When the number of samples is
$10^3$, the number of erroneous edges in the output as depicted by Figure
\ref{fig:outerplanar_output1} is $0.18$. However, as the number of samples
increases to $10^4$, the recovered graphical model in Figure
\ref{fig:outerplanar_output2} is exactly the same as the original graphical
model.

\section{An Example Application: Modeling Correlations of Senator Voting}
\begin{figure*}[ht]
\vspace{-.3cm}
\centering \includegraphics[width=17cm]{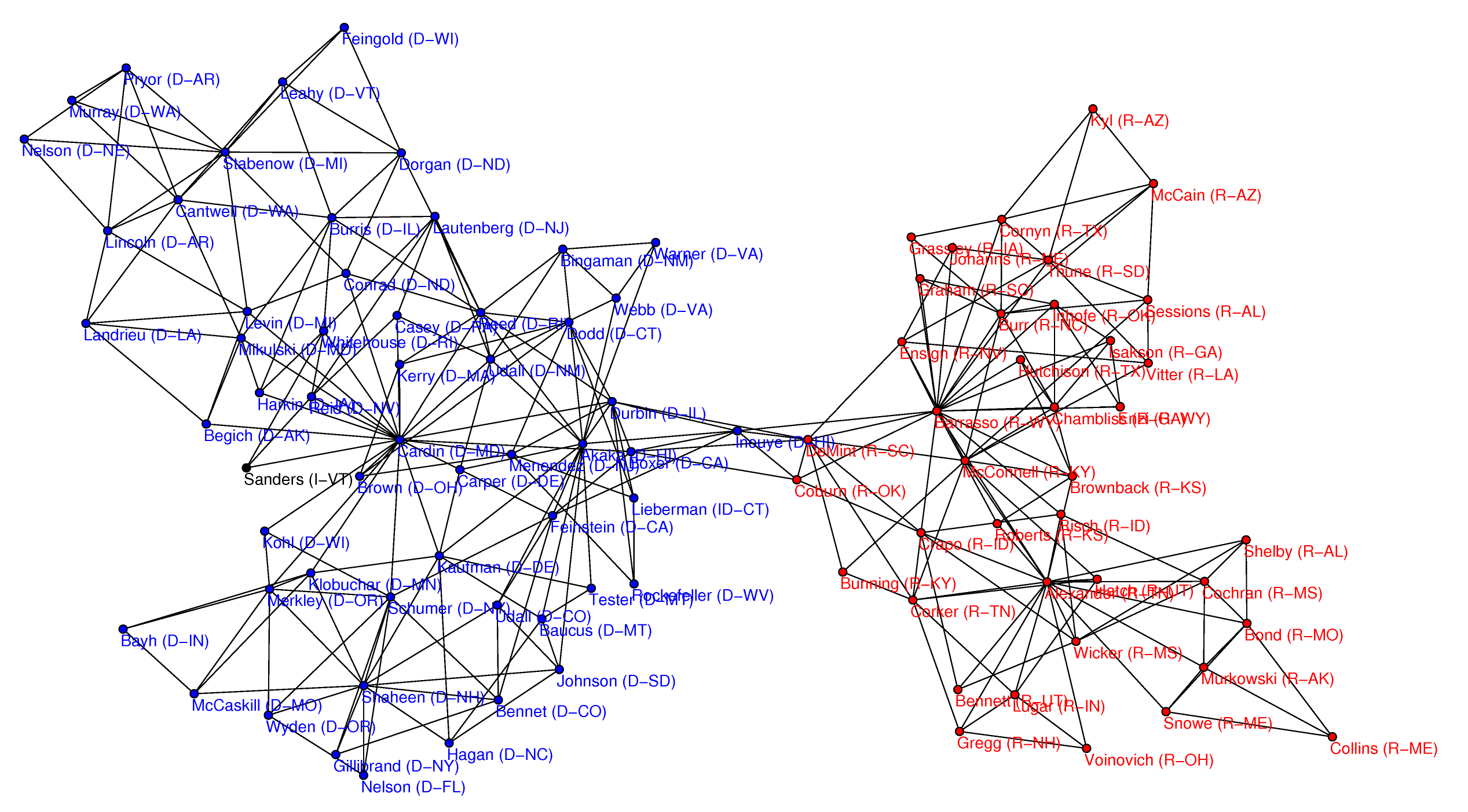}
\label{fig:senatorgraph}
\vspace{-.5cm}
\caption{Graphical model representing the senator voting pattern : The blue nodes represent democrats, the red nodes
represent republicans and the black node represents an independent. The above graphical model conveys many facts that
are already known to us. For instance, the graph shows Sanders with edges only to democrats which makes sense because
he caucuses with democrats. Same is the case with Lieberman. The graph also shows the senate minority leader
McConnell well connected to other republicans though the same is not true of the senate majority leader Reid.
We use the graph drawing algorithm of Kamada and Kawai \cite{kamada}.}
\end{figure*}

In this section, we use our algorithm in an interesting application to model correlations of senator voting following
Banerjee et al. \cite{banerjee}.
We use the senator voting data for the years 2009 and 2010 to calculate the correlations between the voting patterns
of different senators. A \textit{Yea} vote is treated as $+1$ and a \textit{Nay} vote is treated as $-1$. We also
consider non-votes as $-1$, but only consider those senators who voted in atleast $\frac{3}{4}$ of the votes under consideration
to limit bias. We run our algorithm on the correlation data to obtain the maximal planar graph modeling the senator voting
pattern which is presented in Figure \ref{fig:senatorgraph}.

\section{Conclusion}

We have proposed a greedy heuristic to obtain the maximum-likelihood planar
Ising model approximation to a collection of binary random variables with known
pairwise marginals. The algorithm is simple to
implement with the help of known methods for tractable inference in planar Ising
models, efficient methods for planarity testing and embedding of planar
graphs. We have presented simulation results of our algorithm on sample data and
on the senate voting record.
%  and demonstrate its applicability in
% many practical problems.

\paragraph{Future Work} Many directions for further work are suggested by the
methods and results of this paper. Firstly, we know that the greedy algorithm is
not guaranteed to find the best planar graph. Hence, there are several
strategies one might consider to further refine the estimate. One is to allow
the greedy algorithm to also remove previously-added edges which prove to be
less important than some other edge.

It may also be possible to use some more generalized notion of local search,
such as adding/removing multiple edges at a time such as searching the space of
maximal planar graphs by considering ``edge flips'', that is, replacing an edge
by an orthogonal edge connecting opposite vertices of the two adjacent faces.
One could also consider randomized search strategies such as simulated annealing
or genetic programming in the hope of escaping local minima. Another limitation
of our current framework is that it only allows learning planar graphical models
on the set of observed random variables and, moreover, requires that all
variables are observed in each sample.  One could imagine extensions of our
approach to handle missing samples (using the expectation-maximization approach)
or to try to identify hidden variables that were not seen in the data.

\bibliographystyle{is-alpha}
\bibliography{planar}
\appendix
 
\section*{Supplementary Material (Proofs)}

\begin{proof}[\textbf{Proposition $1$}]
%\paragraph{Proposition \ref{prop:zeromeantononzero}}
Let the probability distributions corresponding to $G$ and $\widehat{G}$ be $P$ and $\widehat{P}$ respectively and
the corresponding expectations be $\mathbb{E}$ and $\widehat{\mathbb{E}}$ respectively.
For the partition function, we have that
\begin{equation*}
 \begin{array}{rl}
  \widehat{Z} &= \displaystyle \sum_{x_{\widehat{V}}} \exp\left( \displaystyle \sum_{\{i,j\} \in \widehat{E}} \widehat{\theta}_{ij}x_ix_j\right) \\
  &= \displaystyle \sum_{x_{\widehat{V}}} \exp\left( x_{n+1}\displaystyle \sum_{i \in V} \theta_{i}x_i+\displaystyle \sum_{\{i,j\} \in E} \theta_{ij}x_ix_j\right) \\
  &= \displaystyle \sum_{x_{V}} \exp\left( \displaystyle \sum_{i \in V} \theta_{i}x_i+\displaystyle \sum_{\{i,j\} \in E} \theta_{ij}x_ix_j\right) \\
  &\;\; + \displaystyle \sum_{x_{V}} \exp\left( -\displaystyle \sum_{i \in V} \theta_{i}x_i+\displaystyle \sum_{\{i,j\} \in E} \theta_{ij}x_ix_j\right) \\
  &= 2\displaystyle \sum_{x_{V}} \exp\left( \displaystyle \sum_{i \in V} \theta_{i}x_i+\displaystyle \sum_{\{i,j\} \in E} \theta_{ij}x_ix_j\right) = 2Z
 \end{array}
\end{equation*}
where the fourth equality follows from the symmetry between $-1$ and $1$ in an Ising model.

For the second part, since $\widehat{P}$ is zero-field, we have that
\begin{equation*}
 \widehat{\mathbb{E}}[x_i] = 0 \; \forall \; i \in \widehat{V}
\end{equation*}
Now consider any $\{i,j\} \in E$. If $x_{n+1}$ is fixed to a value of $1$, 
then the model is the same as original on $V$ and we have
\begin{equation*}
 \widehat{\mathbb{E}}[x_ix_j \mid x_{n+1}=1] =  \mathbb{E}[x_ix_j]\; \forall \; \{i,j\} \in E 
\end{equation*}
By symmetry (between $-1$ and $1$) in the model, the same is true for 
$x_{n+1}=-1$ and so we have
\begin{equation*}
  \begin{array}{rl}
    &\widehat{\mathbb{E}}[x_ix_j]\\
    &=\widehat{\mathbb{E}}[x_ix_j \mid x_{n+1}=1]\widehat{P}(x_{n+1}=1)\\
    &\;+ \widehat{\mathbb{E}}[x_ix_j \mid x_{n+1}=-1]\widehat{P}(x_{n+1}=-1)\\
    &= \mathbb{E}[x_ix_j]
  \end{array}
\end{equation*}
Fixing $x_{n+1}$ to a value of $1$, we have
\begin{equation*}
 \widehat{\mathbb{E}}[x_i \mid x_{n+1}=1] =  \mathbb{E}[x_i]\; \forall \; i \in V
\end{equation*}
and by symmetry
\begin{equation*}
 \widehat{\mathbb{E}}[x_i \mid x_{n+1}=-1] =  -\mathbb{E}[x_i]\; \forall \; i \in V
\end{equation*}
Combining the two equations above, we have
\begin{equation*}
  \begin{array}{rl}
    &\widehat{\mathbb{E}}[x_ix_{n+1}]\\
    &=\widehat{\mathbb{E}}[x_i \mid x_{n+1}=1]\widehat{P}(x_{n+1}=1)\\
    &\;+ \widehat{\mathbb{E}}[-x_i \mid x_{n+1}=-1]\widehat{P}(x_{n+1}=-1)\\
    &= \mathbb{E}[x_i]
  \end{array}
\end{equation*}
\end{proof}

\begin{proof}[\textbf{Proposition $2$}]
 From Theorem $1$, we see that the log partition function can be written as
\begin{equation*}
 \Phi(\theta) = n \log 2 + \displaystyle \sum_{\{i,j\}\in E} \log \cosh \theta_{ij} + \frac{1}{2} \log \det(I-AD)
\end{equation*}
where $A$ and $D$ are as given in Theorem $1$. %\ref{thm:partfunc_planarising}.
For the derivatives, we have
\begin{equation*}
\begin{array}{rl}
  \frac{\partial\Phi(\theta)}{\partial \theta_{ij}} &= \tanh \theta_{ij} + \frac{1}{2} \mbox{Tr}\left((I-AD)^{-1} \frac{\partial(I-AD)}{\partial \theta_{ij}}\right)\\
	&= \tanh \theta_{ij} - \frac{1}{2} \mbox{Tr}\left((I-AD)^{-1} A D'_{ij}\right)\\
	&= w_{ij} - \frac{1}{2} (1-w_{ij})^2 \left( S_{ij,ij}+S_{ji,ji}\right)
\end{array}
\end{equation*}
where $D'_{ij}$ is the derivative of the matrix $D$ with respect to $\theta_{ij}$.
The first equality follows from chain rule and the fact that $\nabla K = K^{-1}$ for any matrix $K$. Please refer
\cite{boyd} for details.

For the Hessian, we have
\begin{equation*}
 \begin{array}{rl}
  \frac{\partial^2\Phi(\theta)}{\partial \theta_{ij}^2} &= \frac{1}{Z(\theta)}\frac{\partial^2 Z(\theta)}{\partial \theta_{ij}^2} - \frac{1}{Z(\theta)^2}\left( \frac{\partial Z(\theta)}{\partial \theta_{ij}} \right)^2 \\
	  &= 1 -\mu_{ij}^2
 \end{array}
\end{equation*}
For $\{i,j\}\neq \{k,l\}$, following \cite{boyd}, we have
\begin{equation*}
\begin{array}{rl}
 \frac{\partial^2\Phi(\theta)}{\partial \theta_{ij}\partial \theta_{kl}}
   &= -\frac{1}{2}\mbox{Tr}\left( SD'_{ij}SD'_{kl}\right) \\
   &= -\frac{1}{2} (1-w_{ij}^2) \left( S_{ij,kl}S_{kl,ij}+S_{ji,kl}S_{kl,ji} \right. \\
   &\;\;\;\;\left. +S_{ij,lk}S_{lk,ij}+S_{ji,lk}S_{lk,ji}\right) (1-w_{kl}^2)
\end{array}
\end{equation*}
On the other hand, we also have
\begin{equation*}
 \begin{array}{rl}
  T_{ij,kl} &= e_{ij}^T (I+P) (S\circ S^T) (I+P) e_{kl} \\
	   &= (e_{ij} + e_{ji})^T (S\circ S^T) (e_{kl} + e_{lk}) \\
	   &= (S\circ S^T)_{ij,kl} + (S\circ S^T)_{ij,lk} \\
	   &\;\;\;+ (S\circ S^T)_{ji,kl} + (S\circ S^T)_{ji,lk} \\
	   &= S_{ij,kl} S_{kl,ij}+S_{ji,kl}S_{kl,ji} \\
	   &\;\;\;+S_{ij,lk}S_{lk,ij}+S_{ji,lk}S_{lk,ji}
 \end{array}
\end{equation*}
where $e_{ij}$ is the unit vector with $1$ in the $ij^{\mbox{th}}$ position and $0$ everywhere else.
Using the above two equations, we obtain
\begin{equation*}
 H_{ij,kl} = -\frac{1}{2} (1-w_{ij}^2) T_{ij,kl} (1-w_{kl}^2)
\end{equation*}
\end{proof}

\begin{proof}[\textbf{Proposition $4$}]
The proof follows from the following steps of inequalities.
\begin{equation*}
 \begin{array}{rl}
  D(P,P_G) &= D(P,P_{G+ij}) + D(P_{G+ij},P_G) \\
	   &= D(P,P_{G+ij}) + \\
	   &\;\;\;D(P_{G+ij}(x_i,x_j),P_G(x_i,x_j)) + \\
	   &\;\;\;D(P_{G+ij}(x_{V-ij}),P_G(x_{V-ij})) \\
	   &\geq D(P,P_{G+ij}) + \\
	   &\;\;\;D(P_{G+ij}(x_i,x_j),P_G(x_i,x_j)) + \\
	   &\geq D(P,P_{G+ij}) + \\
	   &\;\;\;D(P(x_i,x_j),P_G(x_i,x_j))
 \end{array}
\end{equation*}
where the first step follows from the Pythagorean law of information projection \cite{amari},
the second step follows from the conditional rule of relative entropy \cite{cover},
the third step follows from the information inequality \cite{cover} and finally
the fourth step follows from the property of information projection to $G+ij$ \cite{wainwright}.
\end{proof}

\end{document}